\documentclass[sigconf]{acmart}

\usepackage{booktabs} 

\usepackage{amsmath}
\usepackage{algorithm}
\usepackage{algorithmic}

\usepackage{subfigure}
\usepackage{multirow}
\usepackage{tabularx}

\copyrightyear{2019}
\acmYear{2019} 
\setcopyright{iw3c2w3}
\acmConference[WWW '19]{Proceedings of the 2019 World Wide Web Conference}{May 13--17, 2019}{San Francisco, CA, USA}
\acmBooktitle{Proceedings of the 2019 World Wide Web Conference (WWW'19), May 13--17, 2019, San Francisco, CA, USA}
\acmPrice{}
\acmDOI{10.1145/3308558.3313658}
\acmISBN{978-1-4503-6674-8/19/05}

\usepackage{balance}

\begin{document}

\title{Efficient Path Prediction for Semi-Supervised and Weakly Supervised Hierarchical Text Classification}

\author{Huiru Xiao}
\affiliation{%
  \institution{Hong Kong University of Science and Technology}
}
\email{hxiaoaf@cse.ust.hk}

\author{Xin Liu}
\affiliation{%
  \institution{Hong Kong University of Science and Technology}
}
\email{xliucr@cse.ust.hk}

\author{Yangqiu Song}
\affiliation{%
  \institution{Hong Kong University of Science and Technology}
}
\email{yqsong@cse.ust.hk}


\begin{abstract}
Hierarchical text classification has many real-world applications.
However, labeling a large number of documents is costly.
In practice, we can use semi-supervised learning or weakly supervised learning (e.g., dataless classification) to reduce the labeling cost.
In this paper, we propose a path cost-sensitive learning algorithm to utilize the structural information and further make use of unlabeled and weakly-labeled data.
We use a generative model to leverage the large amount of unlabeled data and introduce path constraints into the learning algorithm to incorporate the structural information of the class hierarchy. 
The posterior probabilities of both unlabeled and weakly labeled data can be incorporated with path-dependent scores.
Since we put a structure-sensitive cost to the learning algorithm to constrain the classification consistent with the class hierarchy and do not need to reconstruct the feature vectors for different structures, we can significantly reduce the computational cost compared to structural output learning.
Experimental results on two hierarchical text classification benchmarks show that our approach is not only effective but also efficient to handle the semi-supervised and weakly supervised hierarchical text classification.
\end{abstract}

%
%

\ccsdesc[300]{Computing methodologies~Semi-supervised learning settings}
\ccsdesc[100]{Computing methodologies~Classification and regression trees}

\keywords{Hierarchical Text Classification, Semi-Supervised Learning, Weakly Supervised Learning, Structured Prediction}

\maketitle

\section{Introduction}
Text classification has always been an important task, particularly with the vast growth of text data on the Web needed to be classified. The applications include news classification \cite{DaganKR97}, product review classification \cite{PangLV02}, spam detection \cite{McCordC11} and so on. Hierarchical classification (HC) and structured prediction are involved since the classes are usually organized as a hierarchy.
In recent decades, many approaches have been proposed for HC. For example, top-down classification \cite{SunL01} classifies documents at the top layer and then propagates the results to next layer until the leaves. This greedy strategy propagates the classification error along the hierarchy. Contrarily, bottom-up classification \cite{BennettN09} backpropagates the labels from the leaves to the top layer, making the leaves with less training data but sharing some similarities with their parents and siblings may not get well considered and trained. 
Moreover, structural output learning, such as structural perceptron \cite{DBLP:conf/emnlp/Collins02} and structural SVM \cite{DBLP:journals/jmlr/TsochantaridisJHA05}, can leverage the structural information in the class hierarchy well, 
but they need to do Kesler construction \cite{Nilsson65,DudaHart73} where for each sub-structure, the new features are constructed based on the existing features and the class dependencies.
That is why structural output learning usually takes more time to train than top-down and bottom-up approaches.
All the above approaches are supervised methods.
When there are more unlabeled data, it is more challenging if we consider both class dependencies and efficiency of the practical use of hierarchical text classification.

There exist several ways to use the large amount of unlabeled data, among which semi-supervised learning (SSL) \cite{Chapelle:2010:SL:1841234} and weakly supervised learning such as dataless classification \cite{DBLP:conf/aaai/ChangRRS08,song2014dataless} are two representative ways.
An example of SSL is \cite{nigam2006semi}. It uses a mixture multinomial model to estimate the posterior probabilities of unlabeled data, which share the same parameters with the naive Bayes model for the labeled data.
More parameters can be introduced to model the hierarchical structure, causing the model redundant and meanwhile not accurate enough.
As for weakly supervised learning, dataless classification \cite{song2014dataless} uses the semantic similarities between label descriptions and document contents to provide weak labels for documents.
When applying the weak labels, current approaches simply treat each label similarity independently and do not consider the path constraints in the label hierarchy.

To tackle the above problems for semi-supervised and weakly supervised hierarchical text classification, we propose a path cost-sensitive learning algorithm based on a generative model for text.
When estimating the path posterior distribution, the path-dependent scores are incorporated to make the posteriori path-sensitive.
The path-dependent score evaluates how accurate the current model is in terms of classifying a document among the paths in the class hierarchy. Then during inference, classification is constrained to keep the consistency of the hierarchy.
By this mechanism, we develop a simple model with fewer parameters compared with existing approaches while maintaining the consistency property for the class dependencies in the hierarchy.

The contributions of our paper are as follows:
\begin{enumerate}
\item We propose \textbf{a new approach for hierarchical text classification} based on a probabilistic framework. We highlight its meanings in cost-sensitive learning and constraint learning.
\item We show significant improvements on two widely used hierarchical text classification benchmarks and demonstrate our algorithm's \textbf{effectiveness} in \textbf{semi-supervised} and \textbf{weakly-supervised learning} settings.
\item Our approach \textbf{reduces the complexity} of traditional methods. We achieve tens of times speedup while outperforming the state-of-the-art discriminative baseline.
\end{enumerate}

The code and data used in the paper are available at \url{https://github.com/HKUST-KnowComp/PathPredictionForTextClassification}.

\section{Related Work}
There are only a few studies on semi-supervised hierarchical (text) classification~\cite{nigam2006semi,DBLP:conf/wsdm/DalviMC16}, partially because of the difficulty to evaluate the class dependencies for unlabeled data and the time cost of using more complicated algorithms such as structural output learning~\cite{MannM08}. Most semi-supervised hierarchical text classification works were based on EM algorithm introduced by \cite{nigam2006semi}. Some are related with ours (see in Section 2.1), while others are not, e.g., \cite{DBLP:conf/wsdm/DalviMC16} used EM algorithm to deal with incomplete hierarchy problem, which was not the same setting as ours. In this section, we simply start with the review of general hierarchical text classification and then explain the uniqueness and significance of our work.

Hierarchical text classification has been studied for several decades. Flat multi-label classification methods \cite{DBLP:journals/ajiips/TikkB04} ignore the hierarchy, thus poor for HC. Early works \cite{DBLP:conf/icml/KollerS97,DBLP:conf/sigir/DumaisC00,DBLP:journals/sigkdd/LiuYWZCM05} often used ``pachinko-machine models'' which assigned a local classifier at each node and classified documents recursively. Top-down and bottom-up approaches utilize the local classifier ideas, but top-down is a greedy strategy so it may not find optimal solutions, while bottom-up approach does not well consider and train the classes with less training data.

To better exploit the class hierarchy, algorithms particularly designed for trees can assist.
In practice, both generative and discriminative models are used.
In the following, we will review the related work of these two categories.

\subsection{Generative Models}
\cite{nigam2006semi} summarized the text generative model and provided the naive Bayes classifier and Expectation-Maximization (EM) algorithm for flat classification. As for HC, it introduced more parameters to account for the class dependencies. \cite{DBLP:conf/icml/McCallumRMN98} remodeled the framework in another way. They applied shrinkage to smooth parameter estimates using the class hierarchy. \cite{DBLP:conf/dasfaa/CongLWL04} also used the same generative framework but proposed a clustering-based partitioning technique. These generative hierarchical methods can bring some structural information to the model, but they do not make full use of the hierarchy and have difficulties scaling to large hierarchies.

\subsection{Discriminative Models}
Discriminative methods are also popular for HC. 
Orthogonal Transfer \cite{DBLP:conf/icml/XiaoZW11} borrowed the idea of top-down classification where each node had a regularized classifier and each node's normal vector classifying hyperplane was encouraged to be orthogonal to its ancestors'. 
Hierarchical Bayesian Logistic Regression \cite{GopalYBN12} leveraged the hierarchical dependencies by giving the children nodes a prior centered on the parameters of its parents. The idea was further developed in Hierarchically Regularized SVM and Logistic Regression \cite{GopalY13}, where the hierarchical dependencies were incorporated into the parameter regularization structure. 
More recently, the idea of hierarchical regularization has been applied to deep models and also showed some improvements \cite{PengLHLBWS018}.
\cite{CharuvakaR15} simplified the construction of classifier by building a binary classifier on each tree node and providing the cost-sensitive learning (HierCost). All the above approaches are still based on top-down or greedy classification which can result in non-optimal solutions. Another similar work with ours is \cite{DBLP:journals/corr/abs-1709-01062}'s hierarchical loss for classification, which defined the hierarchical loss or win as the weighted sum of the probabilities of the nodes along the path. In contrast to their work, we use the sum of the (weakly) labeled instances along a path as score to perform path cost-sensitive learning. 

To find more theoretically guaranteed solutions, some algorithms were developed based on structural output learning \cite{LaffertyMP01,DBLP:conf/emnlp/Collins02,TaskarGK03,DBLP:journals/jmlr/TsochantaridisJHA05}, which can be proved to be global optimal for HC.
Hierarchical SVM (HSVM) \cite{DBLP:conf/cikm/CaiH04}, one example of structural SVM, generalized SVM to structured data with a path-dependent discriminant function.
In general, when performing structural output learning, Kesler construction is used to construct the feature vectors for comparing different structures \cite{Nilsson65,DudaHart73}, which adds much more computation than top-down or bottom-up classification approaches.

In summary, generative and discriminative models can both be adapted to HC problems. Discriminative models achieve better performance with adequate labeled data~\cite{NgJ01}, especially if a better representation for text can be found, e.g., using deep learning~\cite{PengLHLBWS018}. Whereas generative models have their advantage for handling more uncertainties~\cite{NgJ01} for limited labeled data and under noisy supervision.
Our work is based on a generative model yet has the same parameter size as the flat classification. We find that it significantly boosts the performance of semi-supervised learning and weakly supervised learning as well as reduces the computational cost.

\section{Path Prediction for Hierarchical Classification}
In HC, the classes constitute a hierarchy, denoted as $\mathcal{T}$. $\mathcal{T}$ is a tree whose depth is $d$, with the root node in depth $0$. Then the classes are distributed from depth $1$ to $d$. We suppose that all leaf nodes are in depth $d$. This can always be satisfied by expanding the shallower leaf node (i.e. giving it a child) until it reaches depth $d$.
When evaluating models, these dummy nodes from $\mathcal{T}$ can be easily removed to avoid affecting the performance measure.

Let $\mathcal{C}_1, \dots, \mathcal{C}_d$ be the class sets in depth $1, \dots$, depth $d$ accordingly, with sizes $M_1, \dots, M_d$. To classify a document, we assign labels in each depth, i.e., the document gets $d$ labels $\{c_{j_k}^k: k=1, \dots, d, j_k=1, \dots, M_k\}$. These $\{c_{j_k}^k\}$ form a path in $\mathcal{T}$ if the classification results in each depth are consistent with other depths. We want to maintain the consistency of the hierarchy, therefore we classify the documents by paths instead of by multi-label classes. After assigned a path, the document's classes are the nodes lying in the path. It is similar with structured prediction since a path can be regarded as a structured object, which contains more information than a set of multi-label classes without path constraints.

To sum up, path prediction aims at making use of the structural information in the class hierarchy to train the classifier. Note that the classifier is for paths in the hierarchy instead of classes. The details of path prediction algorithm are given in the next section.

\section{Path Cost-Sensitive Learning}
In this section, we introduce our method which utilizes the structural information to learn the classifier, revealing its meanings in cost-sensitive learning and constraint learning.

\subsection{Path-Generated Probabilistic Framework}
We base our work on a widely-used probabilistic framework, which constructs a generative model for text. In the framework, text data are assumed to be generated from a mixture of multinomial distributions over words. Previous works \cite{nigam2006semi,DBLP:conf/icml/McCallumRMN98} assumed that the mixture components in the generative model have a one-to-one correspondence with the classes. However, in order to perform path prediction, we presume that the mixture components have a one-to-one correspondence with the paths.

Define $\mathcal{P}$ to be the set of all paths which start from the root node and end in some leaf node in the class hierarchy $\mathcal{T}$, so the size of $\mathcal{P}$ equals to that of the leaf nodes $M_d$. Let $\mathcal{V}$ be the vocabulary. Denote $\theta$ as the parameters for the mixture multinomial model. For a document $x_i$ with length $|x_i|$, suppose $x_{it}$ is the word frequency of word $w_t$ in $x_i$, which is the document feature represented by vector space model \cite{liu2012improvement}. Then the generative process runs as following.

First, select a mixture component, or equivalently a path $p_j\in \mathcal{P}$, from $P(p_j|\theta)$ (prior of $p_j$). Next, generate the document $x_i$ by selecting the length $|x_i|$ and picking up $|x_i|$ words from $P(x_i|p_j;\theta)$. According to the law of total probability and the naive Bayes assumption that given the labels, the occurrence times of each word in a document are conditionally independent with its position as well as other words, the probability of generating $x_i$ is
\begin{equation}
\label{equation1}
P(x_i|\theta)\propto P(|x_i|)\sum_{j=1}^{M_d}P(p_j|\theta)\prod_{w_t\in \mathcal{V}}P(w_t|p_j;\theta)^{x_{it}}.
\end{equation}

In general, document lengths are assumed to be independent with classes, thus independent with paths.
So model parameters $\theta$ include the path prior $\theta_{p_j}\equiv P(p_j|\theta)$ and the multinomial distribution over words for each path $\theta_{w_t|p_j}\equiv P(w_t|p_j;\theta)$. 

\subsection{Path-Dependent Scores}
Given a data set $\mathcal{D}=\{(x_i, y_i): x_i\in \mathcal{X}_l\}\cup \{x_i: x_i\in \mathcal{X}_u\}$, consisting of the labeled documents $\mathcal{X}_l$ and the unlabeled documents $\mathcal{X}_u$. We now derive the parameter estimation in a supervised manner.
With only labeled data considered, we maximize $P(\theta|\mathcal{X}_l)$, which can be done by counting the corresponding occurrences of events.
The event counts are usually the hard counts for flat classification. Here we use a path-dependent score to substitute it.

First we define the score of a node in $\mathcal{T}$ for a document. Suppose $x_i\in X_l$, $c_{j_k}^k$ is the $k^{th}$ node in path $p_j$. The node score of $c_{j_k}^k$ for $x_i$, denoted as $S_i(c_{j_k}^k)$, indicates the label of $x_i$. When $x_i$ is labeled with the ground truth labels, $S_i(c_{j_k}^k)=1$ if and only if $c_{j_k}^k$ is one of $x_i$'s labels. We also consider the weakly supervised case. In \cite{song2014dataless}'s dataless text classification, for $x_i\in \mathcal{X}_l$, it is weakly labeled by the semantical similarities with classes. We assign value $1$ to $S_i(c_{j_k}^k)$ if $x_i$ has the largest similarity with $c_{j_k}^k$ among all classes in depth $k$ and 0 otherwise. 

Next we introduce the path score. For $x_i\in X_l$, the score of path $p_j$, denoted as $S_{ij}$, is the sum of the nodes' scores $S_i(c_{j_k}^k)$ lying in $p_j$ except the root node since it makes no sense for classification.
\begin{equation}
\label{equation2}
S_{ij}\equiv S_i(p_j)=\sum_{k=1}^dS_i(c_{j_k}^k).
\end{equation}

\begin{figure}[t]
	\centering
	\includegraphics[width=0.4\textwidth]{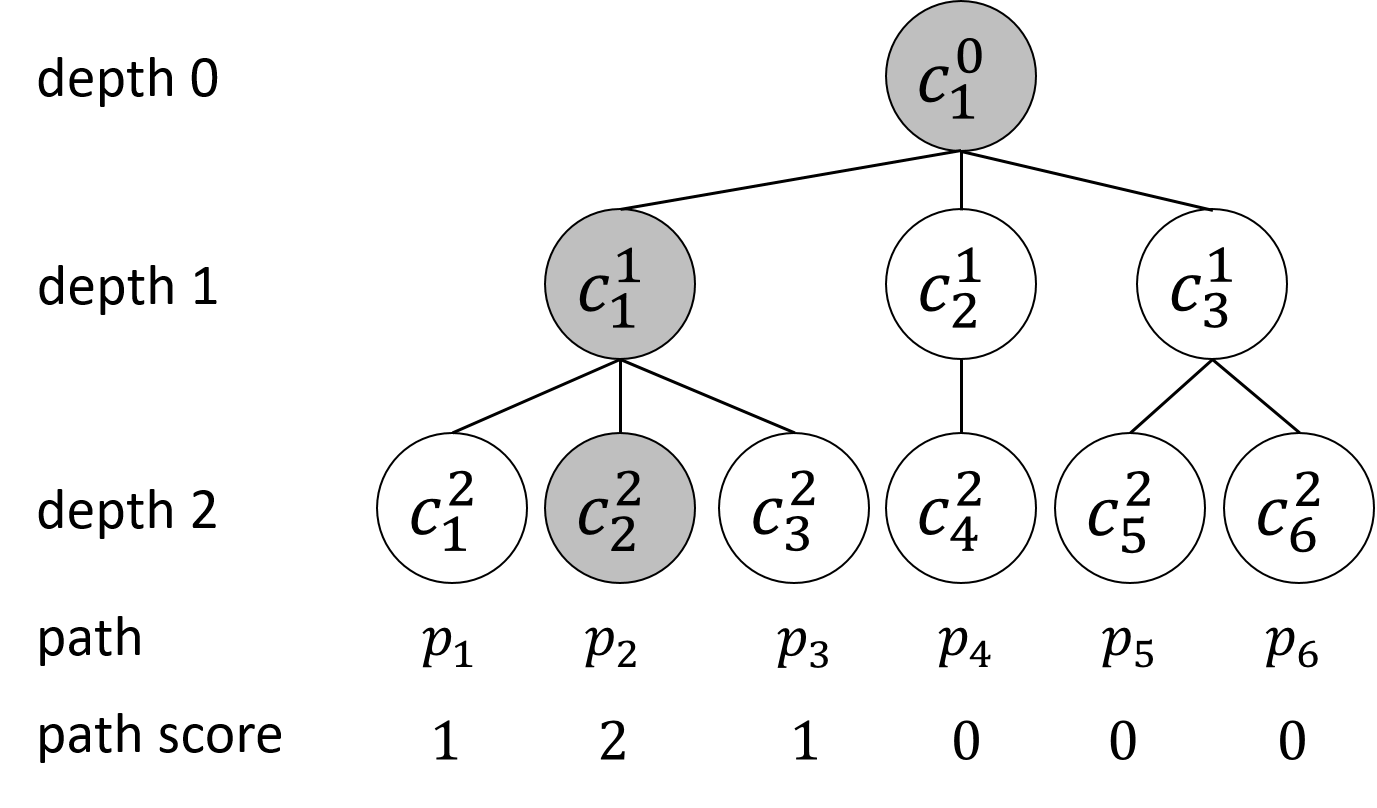}
	\vspace{-0.1in}
	\caption{An example of path scores. Suppose $x_i$ is labeled as $\{c_1^1, c_2^2\}$. $S_{ij}$ for $j=1,\dots,6$ is shown in the figure.}
	\vspace{-0.1in}
    \label{fig:pathscore}
\end{figure}

Take the hierarchy in Figure \ref{fig:pathscore} as an example. $x_i$ is labeled as $\{c_1^1, c_2^2\}$, then $S_{i2}=S_i(c_1^1)+S_i(c_2^2)=2$, $S_{i1}=S_{i3}=1$, while other paths score 0. If $x_i$ is weakly labeled by the similarities, then we label it with the classes having the maximum similarity in each depth and obtain the path scores in the same way. 

\subsection{Path Cost-Sensitive Naive Bayes Classifier}
While doing the empirical counts, the Laplace smoothing is often applied by adding one count to each event to avoid zero probability and shrink the estimator. 
Combining the event counts (i.e. the path scores) and the smoothing term, the parameter estimates are: 
\begin{align}
\label{equation3}
&\hat{\theta}_{p_j}\equiv P(p_j|\hat{\theta})=\frac{1+\sum_{x_i\in \mathcal{X}_l}S_{ij}}{M_d+\sum_{k=1}^{M_d}\sum_{x_i\in \mathcal{X}_l}S_{ik}},\\
\label{equation4}
&\hat{\theta}_{w_t|p_j}\equiv P(w_t|p_j;\hat{\theta})=\frac{1+\sum_{x_i\in \mathcal{X}_l}S_{ij}x_{it}}{|\mathcal{V}|+\sum_{s=1}^{|\mathcal{V}|}\sum_{x_i\in \mathcal{X}_l}S_{ij}x_{is}}.
\end{align}

There are two aspects of using the path scores as event counts:
\begin{enumerate}
\item Cost-sensitive performance measures are considered since different data samples are given different weights. In Figure \ref{fig:pathscore}, $x_i$ is counted twice for $p_2$, once in $c_1^1$ and once in $c_2^2$, thus obtaining more weights. $p_1$ and $p_3$ are not right paths for $x_i$, but they still classify correctly in depth $1$, thus get one count, less than $p_2$ but larger than other paths who have no correct labels at all. This path \textbf{cost-sensitive learning} behavior helps the model to maintain structural information.
\item The path scores function as the measuring indicators of paths, capacitating the model to classify the documents by paths. The path prediction actually puts constraints on the classifier, where the prediction results must be consistent with the class hierarchy. Furthermore, the \textbf{constraint learning} reduces the search space and improves efficiency.
\end{enumerate}

After estimating $\theta$ from $\mathcal{X}_l$, for any test document $x_i$, the posterior probability distribution can be obtained by Bayes' rule:
\begin{align}
\label{equation5}
P(y_i=p_j|x_i;\hat{\theta})&=\frac{P(p_j|\hat{\theta})P(x_i|p_j;\hat{\theta})}{P(x_i|\hat{\theta})} \nonumber \\
&=\frac{\hat{\theta}_{p_j}\prod_{w_t\in \mathcal{V}}(\hat{\theta}_{w_t|p_j})^{x_{it}}}{\sum_{k=1}^{M_d}\hat{\theta}_{p_k}\prod_{w_t\in \mathcal{V}}(\hat{\theta}_{w_t|p_k})^{x_{it}}}.
\end{align}
Then $x_i$ will be classified into $\arg\max_j P(y_i=p_j|x_i;\hat{\theta})$. 

The path cost-sensitive naive Bayes classifier (PCNB) for the generative model are introduced above. Next we will present the semi-supervised path cost-sensitive learning algorithm.

\subsection{Semi-Supervised Path Cost-Sensitive Learning}
Until now, only the labeled data are used during training, but we want to make use of the unlabeled data to ameliorate the classifier. We follow \cite{nigam2006semi} to apply EM technique for SSL.

When the initial parameters are given, the posterior probabilities of $\mathcal{X}_u$, computed through Eq. \eqref{equation5}, can act as the path score $S_{ij}$ for $x_i\in X_u$. 
Combining the labeled and unlabeled data together, the parameter estimates are changed into
\begin{align}
\label{equation6}
&\hat{\theta}_{p_j}=\frac{1+\sum_{x_i\in \mathcal{X}_l\cup \mathcal{X}_u}S_{ij}}{M_d+\sum_{k=1}^{M_d}\sum_{x_i\in \mathcal{X}_l\cup \mathcal{X}_u}S_{ik}},\\
\label{equation7}
&\hat{\theta}_{w_t|p_j}=\frac{1+\sum_{x_i\in \mathcal{X}_l\cup \mathcal{X}_u}S_{ij}x_{it}}{|\mathcal{V}|+\sum_{s=1}^{|\mathcal{V}|}\sum_{x_i\in \mathcal{X}_l\cup \mathcal{X}_u}S_{ij}x_{is}}.
\end{align}

Note that the numerical value of $S_{ij}$ for $x_i\in X_u$ ranges in $[0, 1]$ since it is the posterior probability. Therefore, the unlabeled data weight less than the labeled data while estimating the parameters. It is reasonable because the labeled data are more authentic than the inference results of unlabeled data, especially in the early iterations where the model does not reach convergence. 

The new $\theta$ obtained via Eqs. \eqref{equation6} and \eqref{equation7} are then used to compute the posterior probabilities of $\mathcal{X}_u$ again, which in turn update $\theta$. The iterative process keeps maximizing the likelihood of the dataset $P(\theta|\mathcal{D})$, equivalent to maximizing the log likelihood:
\begin{align}
\label{equation8}
&l(\theta|X,Y)\propto log(P(\theta)P(\mathcal{D}|\theta))
=log(P(\theta)) \nonumber \\
+&\sum_{x_i\in \mathcal{X}_l}log(P(y_i=p_j|\theta)P(x_i|y_i=p_j;\theta)) \nonumber \\
+&\sum_{x_i\in \mathcal{X}_u}log\sum_{p_j\in \mathcal{P}}P(p_j|\theta)P(x_i|p_j;\theta).
\end{align}

Refer to \cite{dempster1977maximum}, the convergence of EM can be guaranteed, but it reaches some local maxima. To enable the algorithm to find good local maxima, we initialize $\theta$ with those obtained through the naive Bayes classifier on $\mathcal{X}_l$. Algorithm \ref{algorithm2} presents the EM algorithm for the path cost-sensitive classification (PCEM).

\begin{algorithm}[t]
\caption{Path Cost-Sensitive Algorithm with EM}
\label{algorithm2}
\begin{algorithmic} 
\REQUIRE Training data $\mathcal{D}=\mathcal{X}_l\cup \mathcal{X}_u$, test data $\mathcal{D}_{test}=\mathcal{X}_{test}$, class hierarchy $\mathcal{T}$.
\STATE \textbf{Training:}
\begin{itemize}
\STATE Compute path scores for $\mathcal{X}_l$, using Eq. \eqref{equation2}.
\STATE Initialize the classifier $\hat{\theta}$ with the path-constrained naive Bayes classifier (Section 4.3).
\WHILE {$l(\hat{\theta})$ (Eq. \eqref{equation8}) not converged}
\STATE \textbf{E-step}: use the current $\hat{\theta}$ to compute the path scores of $\mathcal{X}_u$ with Eq. \eqref{equation5}.
\STATE \textbf{M-step}: use the path scores of $\mathcal{X}_l$ and $\mathcal{X}_u$ to re-estimate $\hat{\theta}$, with Eqs. \eqref{equation6} and \eqref{equation7}.
\ENDWHILE
\end{itemize}
\STATE \textbf{Testing:}
\begin{itemize}
\STATE Compute the posterior probabilities for $\mathcal{X}_{test}$ with Eq. \eqref{equation5}. Classify $\mathcal{X}_{test}$ by selecting $\hat{j}=\arg\max_j P(y_i=p_j|x_i;\hat{\theta})$ for $x_i\in \mathcal{X}_{test}, p_j\in \mathcal{P}$. The classes of $x_i$ are nodes in $p_{\hat{j}}$.
\end{itemize}
\ENSURE The classifier $\hat{\theta}$ and classification results for $\mathcal{D}_{test}$.
\end{algorithmic}
\end{algorithm}

\section{Experiments}
For empirical evaluation of effectiveness and efficiency of our approach, we design experiments on semi-supervised and weakly supervised hierarchical text classification tasks, compared to the representative and the state-of-the-art baselines.
\begin{figure*}[t]
	\centering
	\includegraphics[width=0.4\textwidth]{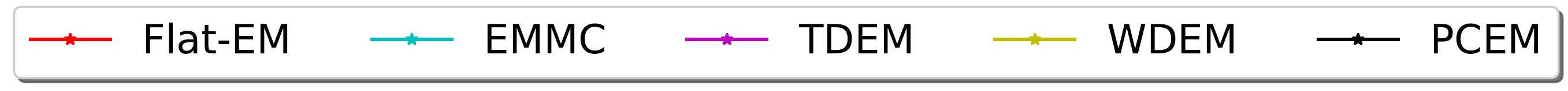}\\
	\vspace{-0.1in}
    \subfigure[Evaluation Results on 20NG]{\label{fig:figs/20ng_labeled}
		\includegraphics[height=4cm, width=0.48\textwidth]{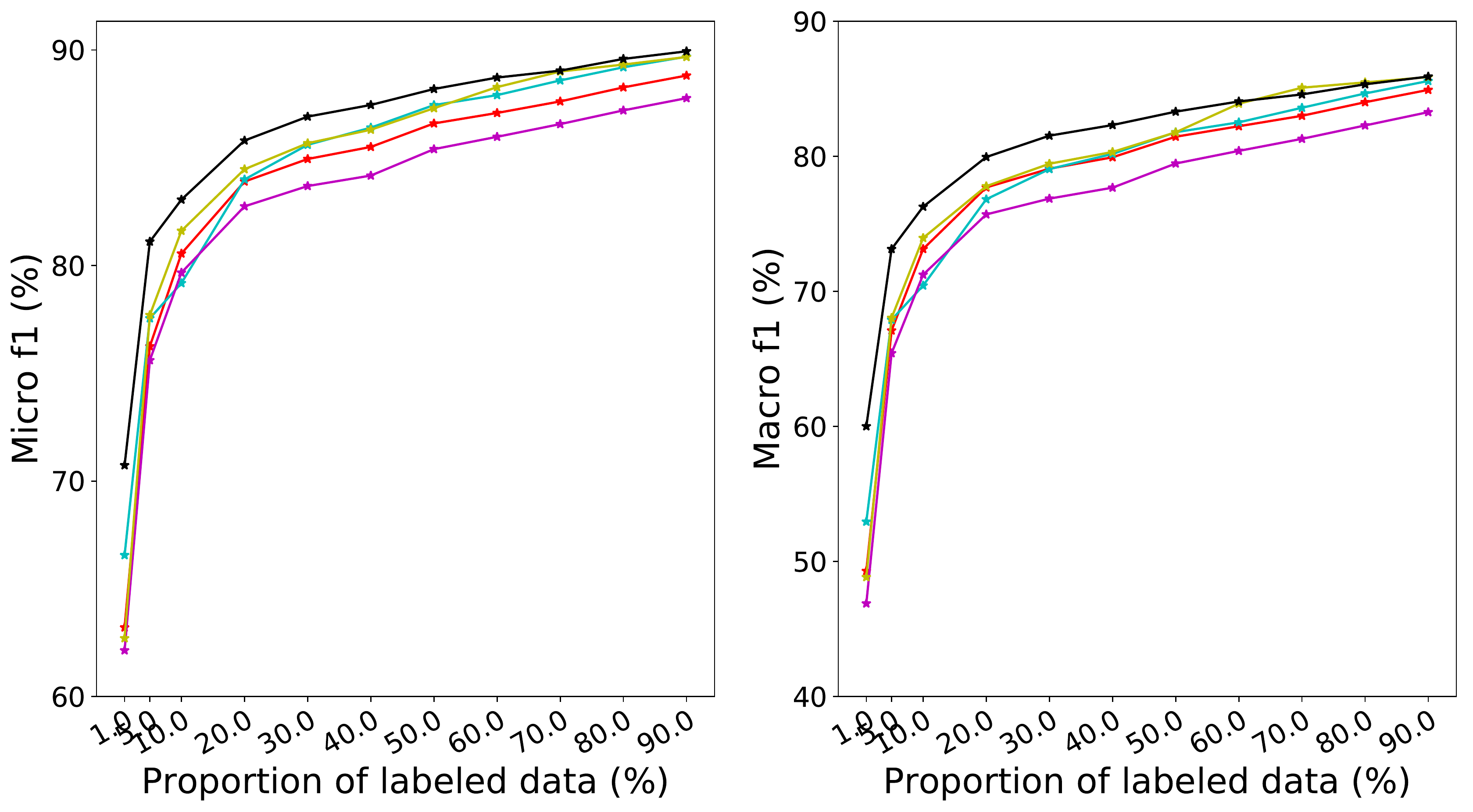}
	}
	\subfigure[Evaluation Results on RCV1]{\label{fig:figs/rcv1_labeled}
		\includegraphics[height=4cm, width=0.48\textwidth]{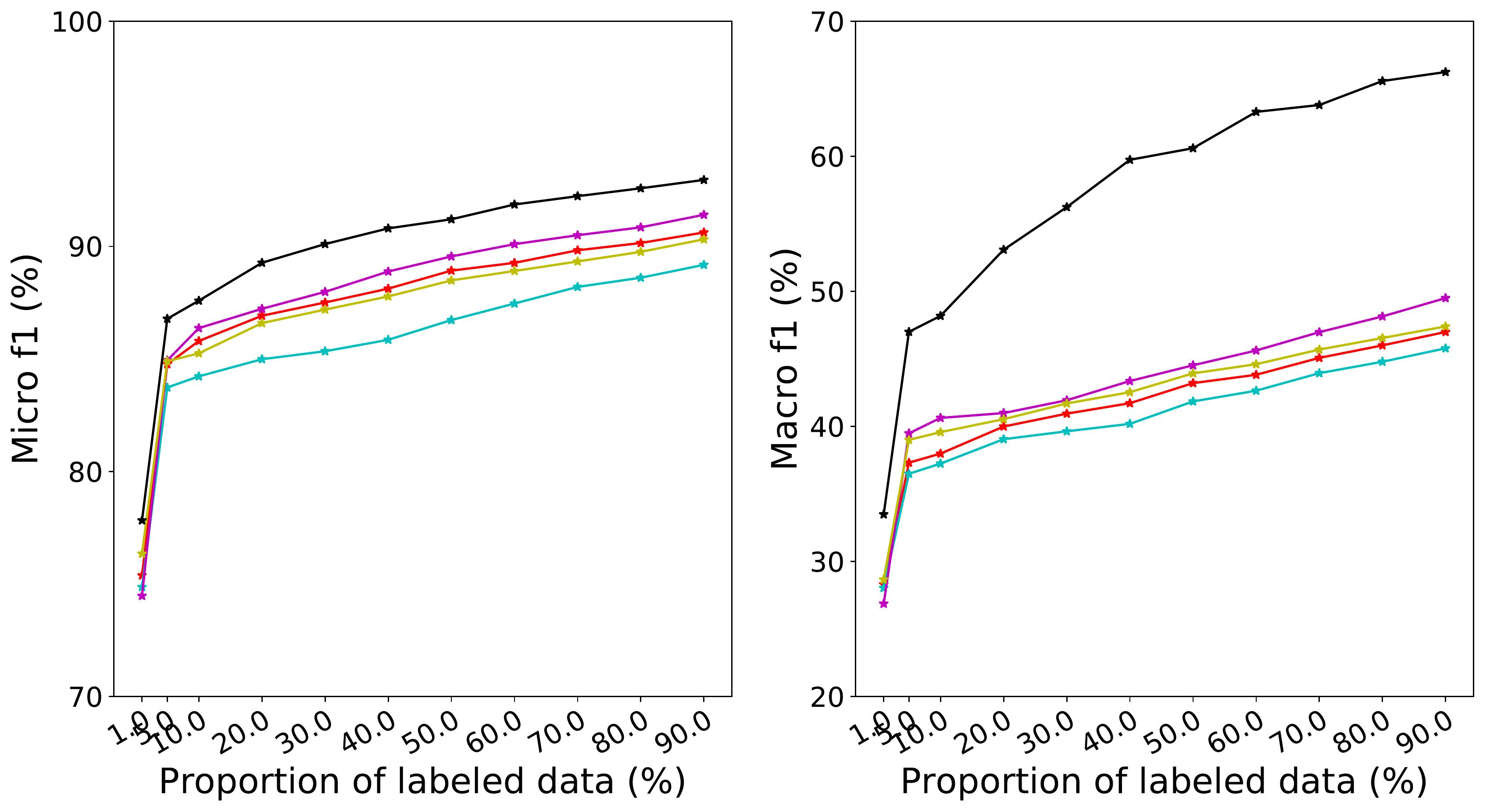}
	}
    \vspace{-0.1in}
	\caption{Experimental results on \textbf{semi-supervised} classification. (a)-(d) shows the micro-$F_1$ score and macro-$F_1$ score on 20NG and RCV1. The horizontal axis is label rate (\%). The vertical axis represents the $F_1$ scores (\%).}
    \label{fig:semisupervised}
\end{figure*}

\begin{figure*}[t]
	\centering
	\includegraphics[width=0.4\textwidth]{figs/legend}\\
	\vspace{-0.1in}
    \subfigure[Evaluation Results on 20NG]{\label{fig:figs/20ng_dataless}
		\includegraphics[height=4cm, width=0.48\textwidth]{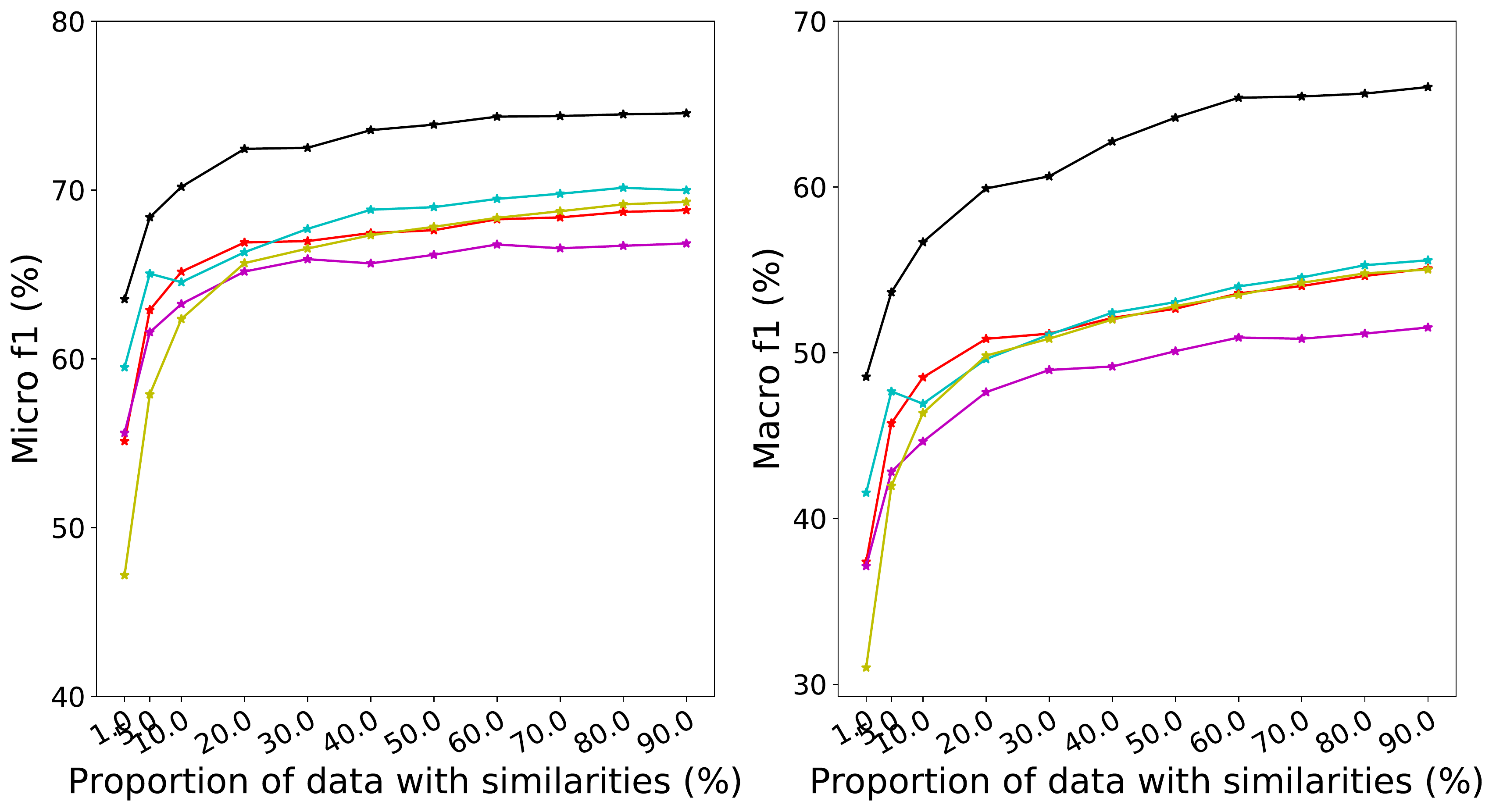}
	}
	\subfigure[Evaluation Results on RCV1]{\label{fig:figs/rcv1_dataless}
		\includegraphics[height=4cm, width=0.48\textwidth]{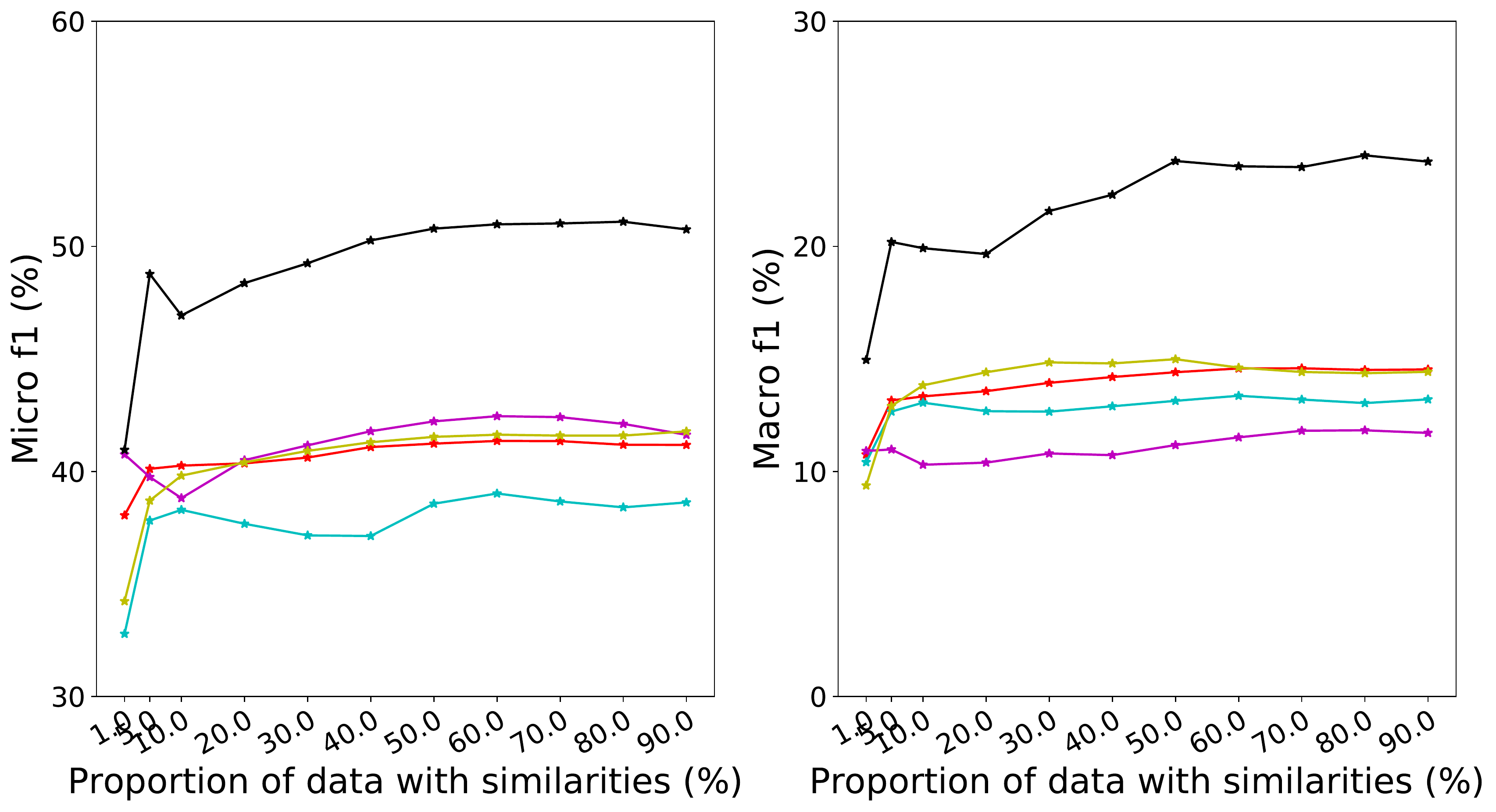}
	}
	\vspace{-0.1in}
	\caption{Experimental results on \textbf{weakly-supervised} classification. (a)-(d) shows the micro-$F_1$ scores and macro-$F_1$ scores on dataless 20NG and RCV1. The horizontal axis is the proportion of data with semantical similarities (\%) in training set. The vertical axis represents the $F_1$ scores (\%).}
    \vspace{-0.1in}
    \label{fig:dataless}
\end{figure*}

\subsection{Experimental Design}

\subsubsection{Datasets}
We use two datasets, both of which have semi-supervised and weakly-supervised version. The statistics are listed in Table \ref{table:statistics}. 
\begin{table}[t]
    \centering
    {
    \small
    \begin{tabularx}{\linewidth}{l|c|c|c|c|c|c}
        \toprule
         & \#Training & \#Test & \#Features & \#Leaves & \#Nodes & Depth \\
        \midrule
        20NG & 15,077 & 3,769 & 103,363 & 20 & 27 &2\\
        RCV1 & 6,395 & 1,733 & 26,888 & 35 & 56 & 3\\
        \bottomrule
    \end{tabularx}
    }
    \caption{Statistics of the datasets.}
    \vspace{-0.2in}
    \label{table:statistics}
\end{table}

\begin{itemize}  
\item \textbf{20NG}\footnote{\url{http://qwone.com/~jason/20Newsgroups/}} \cite{Lang95} 20 Newsgroups is a widely-used text classification dataset. To experiment on weakly-supervised setting and compare with semi-supervised baselines, we use dataless 20NG provided in \cite{song2014dataless}. 
\item \textbf{RCV1} \cite{lewis2004rcv1} RCV1 is a collection of manually labeled Reuters News from 1996-1997. We also use the subset of data provided in \cite{song2014dataless}, which consists of 8,668 single-labeled news documents on the dataless setting.
\end{itemize}

\subsubsection{Baselines}
We compare our path cost-sensitive algorithms (\textbf{PCNB} and \textbf{PCEM}) with the following baselines:
\begin{itemize} 
\item \textbf{Generative baselines}
\begin{enumerate}
  \item Flat naive Bayes classifier (\textbf{Flat-NB}) and \textbf{Flat-EM} algorithm: the flat classifiers introduced in \cite{nigam2006semi}.
  \item Naive Bayes classifier with multiple components (\textbf{NBMC}) and \textbf{EMMC}: a more expressive model proposed by \cite{nigam2006semi}.
  \item Top-down naive Bayes classifier (\textbf{TDNB}) and \textbf{TDEM}: the classifiers run in the top-down way.
  \item Win-driven naive Bayes classifier (\textbf{WDNB}) and \textbf{WDEM}: the modified hierarchical loss for classification \cite{DBLP:journals/corr/abs-1709-01062}.
\end{enumerate}
\item \textbf{Discriminative baselines}
\begin{enumerate}
  \item Logistic regression (\textbf{LR}) and \textbf{SVM}: two classical discriminative methods. Our experiments use the LibLinear\footnote{ \url{https://www.csie.ntu.edu.tw/}} \cite{fan2008liblinear} to train corresponding models and test. During the experiments, we found that dual solvers were much faster and even better in performance than primal solvers, so we chose dual solvers.
  \item \textbf{HierCost}\footnote{\url{https://cs.gmu.edu/~mlbio/HierCost/}} \cite{CharuvakaR15}: the state-of-the-art discriminative method for hierarchical text classification.
\end{enumerate}
\end{itemize}
\begin{table*}[t]
    \centering
    \setlength\tabcolsep{5pt}
    {
    \begin{tabular}{l|c|c|c|c|c|c|c|c}
        \toprule
        & \multicolumn{4}{c|}{20NG} & \multicolumn{4}{c}{RCV1} \\
        & \multicolumn{2}{c|}{labeled} & \multicolumn{2}{c|}{dataless} & \multicolumn{2}{c|}{labeled} & \multicolumn{2}{c}{dataless} \\
        \midrule
      	LR              & \ddag{52.02} & \ddag{42.41} & \ddag{44.16} & \ddag{31.51} & \ddag{69.59} & \dag{24.43} & \dag{33.54} & \ddag{9.00}\\
        SVM             & \ddag{48.33} & \ddag{39.73} & \ddag{41.70} & \ddag{30.24} & \ddag{68.78} & \dag{23.97} & \dag{34.15} & \dag{9.72}  \\
        HierCost        & \ddag{48.12} & \ddag{40.89} & \ddag{43.26} & \ddag{32.30} & \ddag{69.22} & \dag{24.98} & \dag{31.07} & \ddag{8.84} \\
        \midrule
        NB              & \ddag{53.39} & \ddag{39.94} & \ddag{47.29} & \ddag{30.67} & \ddag{70.68} & \dag{24.48} & \dag{33.29} & \ddag{8.39} \\
        NBMC            & \ddag{46.99} & \ddag{38.02} & \ddag{43.24} & \ddag{28.82} & \dag{69.84} & \dag{23.52} & \dag{28.91} & \ddag{6.91} \\
        TDNB            & \ddag{55.50} & \ddag{42.16} & \ddag{48.06} & \ddag{31.02} & \ddag{70.37} & \dag{24.65} & 33.67 & \dag{8.40} \\
        WDNB            & \ddag{53.66} & \ddag{41.53} & \ddag{47.19} & \ddag{31.02} & \ddag{70.89} & \dag{25.04} & 34.24 & \ddag{9.38}  \\
        \textbf{PCNB}   & \ddag{58.33} & \ddag{48.04} & \ddag{52.14} & \dag{38.50} & \dag{73.63} & 29.95 & 37.06 & 12.47 \\
        \midrule
        EM              & \dag{63.21} & \dag{49.30} & \dag{55.13} & \dag{37.40} & 75.38 & 28.32 & 38.05 & \dag{10.76} \\
        EMMC            & \dag{66.56} & \dag{52.95} & 59.50 & \dag{41.56} & 74.86 & 28.04 & 32.79 & \dag{10.42}  \\
        TDEM            & \dag{62.14} & \ddag{46.89} & \dag{55.62} & \dag{37.14} & \dag{74.48} & \dag{26.88} & 40.76 & \dag{10.91}  \\
        WDEM            & \dag{62.71} & \ddag{48.85} & \ddag{47.19} & \ddag{31.02} & 76.35 & 28.66 & 34.24 & \ddag{9.38} \\
        \textbf{PCEM}   & \textbf{70.73} & \textbf{60.02} & \textbf{63.54} & \textbf{48.56} & \textbf{77.83} & \textbf{33.49} & \textbf{40.96} & \textbf{14.96}  \\
        \bottomrule
    \end{tabular}
    }
    \caption{$F_1$ scores (\%) on 1\% "labeled" data of each dataset. Under each dataset, the two columns are Micro-$F_1$ (left) and Macro-$F_1$ (right). The best results are shown in boldface. The statistical significance metrics are marked with either \dag { } if p-values $< 0.05$ or \ddag { } if p-values $< 0.001$.}
     \vspace{-0.2in}
    \label{table:supervisedresults_overall}
\end{table*}

\subsubsection{Evaluation Metrics}
We use $F_1$ scores \cite{DBLP:journals/ir/Yang99} to evaluate the performances of all methods. Denote $TP_i$, $FP_i$, $FN_i$ as the instance numbers of true-positive, false-positive and false negative for class $c_i$. Let $\mathcal{C}$ be the set of all classes except the root node. Two conventional $F_1$ scores are defined as:
\begin{itemize}
\item $\textbf{Micro-$F_1$}=\frac{2PR}{P+R}$,\\
where $P=\frac{\sum_{c_i\in \mathcal{C}}TP_i}{\sum_{c_i\in \mathcal{C}}(TP_i+FP_i)}$ is the averaged precision and $R=\frac{\sum_{c_i\in \mathcal{C}}TP_i}{\sum_{c_i\in \mathcal{C}}(TP_i+FN_i)}$ is the averaged recall.
\item $\textbf{Macro-$F_1$}=\frac{1}{|\mathcal{C}|}\sum_{c_i\in \mathcal{C}}\frac{2P_iR_i}{P_i+R_i}$,\\
where $P_i=\frac{TP_i}{TP_i+FP_i}$ and $R_i=\frac{TP_i}{TP_i+FN_i}$ are the precision and the recall for $c_i$.
\end{itemize}

For the two $F_1$ scores, we measure the overall performance of all classes in the hierarchy in our experiments. 

\subsection{Results}
To evaluate our algorithms, we compare our algorithms with the baselines in semi-supervised and weakly supervised hierarchical text classification. Results on all datasets under $1\%$ label rate are summarized in Table \ref{table:supervisedresults_overall}, where $1\%$ label rate means there are $1\%$ data in the training set are labeled or weakly-labeled, which is a common setting for semi-supervised text classification. To show that our approach (PCEM) indeed levareges unlabeled data and weakly labeled data well, we present the results under different label rates compared with other EM methods in Figure \ref{fig:semisupervised} and \ref{fig:dataless}. For each experiment, we randomly split the training data into labeled and unlabeled according to the label rate, then run experiments using the splitted training data. The running is executed for 5 times and the mean $F_1$ scores are calculated. Next we will analyze the results. Time efficiency will also be discussed.

\subsubsection{Semi-Supervised Classification with True Labels}
Table \ref{table:supervisedresults_overall} shows that when the training data are partly labeled with the ground truth labels, PCEM has remarkable superiority over other methods all the time. The discriminative baselines do not have their advantages on the semi-supervised and weakly supervised settings. When compared with generative baselines, our approaches, either naive Bayes (PCNB) or semi-supervised (PCEM), are the best among the corresponding methods. It demonstrates that our algorithms makes good use of the structural information to improve the hierarchical classification.

As expected, EM approaches outperform the corresponding naive Bayes classifier under $1\%$ label rate, which reveals the benefits from the unlabeled data. However, we also noticed that EM may be surpassed by NB when the label rate gets larger. That is related with whether the ratio between labeled and unlabeled data is suitable for SSL, as well as the bias of unlabeled data. This issue has been discussed in previous works \cite{DBLP:journals/jmlr/Fox-RobertsR14}. 

To see the performance in SSL, we compare PCEM with other EM methods in Figure \ref{fig:semisupervised}. The label rate ranges in $[1.0, 90.0](\%)$. We find that PCEM outperforms others steadily. Other hierarchical EM methods are close to Flat-EM, showing that they takes little advantage of the class hierarchy.
The results reveal the effectiveness of PCEM under all label rates for semi-supervised classification.

\subsubsection{Weakly-Supervised Classification on Dataless Setting}
We also make a comparison with the baselines on dataless text classification. The experimental setting is the same as the semi-supervised classification, except that the training documents do not have labels. Instead, some of them are `labeled' as classes with the maximal semantical similarities. We use the dataless 20NG and RCV1 datasets provided by \cite{song2014dataless}. Results are presented in Table \ref{table:supervisedresults_overall} and Figure \ref{fig:dataless}. 

We find the consistent results with the semi-supervised setting. PCEM can always beat the baselines with significant improvements. PCNB is also better than other NB methods. It is worth noting that the gaps between our algorithms (PCNB and PCEM) and the baselines are bigger than those in the semi-supervised setting. We think the reason is that for this weakly-labeled dataset, the similarities can be seen as noisy labels for documents. In this noisy circumstance, our path cost-sensitive learning algorithm with the probabilistic framework is pretty good at making use of the structural information and features of unlabeled data to recover the true generative distribution.

\subsubsection{Efficiency Comparison}
Time complexity is also under consideration to evaluate our algorithms. PCNB is highly efficient, faster than all of the other methods except Flat-NB and even competitive with Flat-NB. PCEM is slightly slower than LR and SVM, but that is because EM methods leverage the unlabeled data, which cannot be used by discriminative methods. The trade-off is acceptable, especially considering the excellent performance of PCEM.
Furthermore, PCEM also achieves tens of times speedup compared to HierCost.

\section{Conclusions and Future Work}
We present an effective and efficient approach for hierarchical text classification. Our path cost-sensitive learning algorithm alters the traditional generative model of text with a path-generated model to constrain the classifier by the class hierarchy. We show that our algorithm outperforms other baselines on semi-supervised learning and weakly supervised learning.
In addition, our model has the potential of extension to other models, not limited to the generative one, if the path-dependent scores are incorporated appropriately.
For the possible future work, we will convert the current framework into a discriminative learning framework following~\cite{DBLP:conf/emnlp/Collins02} and apply deep neural models to learn a better representation for text~\cite{DBLP:journals/corr/abs-1812-11270,DBLP:conf/cikm/MengSZ018}. 
Discrimative framework will further improve the learning when there are more labeled data and deep neural models are more powerful to handle different kinds of weak supervision.

\section{Acknowledgement}
This paper was supported by HKUST-WeChat WHAT Lab and the Early Career Scheme (ECS, No. 26206717) from Research Grants Council in Hong Kong.

\bibliographystyle{ACM-Reference-Format}
\balance 
\bibliography{Bibliography-File}


\begin{thebibliography}{40}


\ifx \showCODEN    \undefined \def \showCODEN     #1{\unskip}     \fi
\ifx \showDOI      \undefined \def \showDOI       #1{#1}\fi
\ifx \showISBNx    \undefined \def \showISBNx     #1{\unskip}     \fi
\ifx \showISBNxiii \undefined \def \showISBNxiii  #1{\unskip}     \fi
\ifx \showISSN     \undefined \def \showISSN      #1{\unskip}     \fi
\ifx \showLCCN     \undefined \def \showLCCN      #1{\unskip}     \fi
\ifx \shownote     \undefined \def \shownote      #1{#1}          \fi
\ifx \showarticletitle \undefined \def \showarticletitle #1{#1}   \fi
\ifx \showURL      \undefined \def \showURL       {\relax}        \fi
\providecommand\bibfield[2]{#2}
\providecommand\bibinfo[2]{#2}
\providecommand\natexlab[1]{#1}
\providecommand\showeprint[2][]{arXiv:#2}

\bibitem[\protect\citeauthoryear{Bennett and Nguyen}{Bennett and
  Nguyen}{2009}]%
        {BennettN09}
\bibfield{author}{\bibinfo{person}{Paul~N. Bennett} {and} \bibinfo{person}{Nam
  Nguyen}.} \bibinfo{year}{2009}\natexlab{}.
\newblock \showarticletitle{Refined experts: improving classification in large
  taxonomies}. In \bibinfo{booktitle}{\emph{{SIGIR}}}.
  \bibinfo{publisher}{{ACM}}, \bibinfo{pages}{11--18}.
\newblock


\bibitem[\protect\citeauthoryear{Cai and Hofmann}{Cai and Hofmann}{2004}]%
        {DBLP:conf/cikm/CaiH04}
\bibfield{author}{\bibinfo{person}{Lijuan Cai} {and} \bibinfo{person}{Thomas
  Hofmann}.} \bibinfo{year}{2004}\natexlab{}.
\newblock \showarticletitle{Hierarchical document categorization with support
  vector machines}. In \bibinfo{booktitle}{\emph{{CIKM}}}.
  \bibinfo{publisher}{{ACM}}, \bibinfo{pages}{78--87}.
\newblock


\bibitem[\protect\citeauthoryear{Chang, Ratinov, Roth, and Srikumar}{Chang
  et~al\mbox{.}}{2008}]%
        {DBLP:conf/aaai/ChangRRS08}
\bibfield{author}{\bibinfo{person}{Ming{-}Wei Chang},
  \bibinfo{person}{Lev{-}Arie Ratinov}, \bibinfo{person}{Dan Roth}, {and}
  \bibinfo{person}{Vivek Srikumar}.} \bibinfo{year}{2008}\natexlab{}.
\newblock \showarticletitle{Importance of Semantic Representation: Dataless
  Classification}. In \bibinfo{booktitle}{\emph{{AAAI}}}.
  \bibinfo{publisher}{{AAAI} Press}, \bibinfo{pages}{830--835}.
\newblock


\bibitem[\protect\citeauthoryear{Chapelle, Schlkopf, and Zien}{Chapelle
  et~al\mbox{.}}{2010}]%
        {Chapelle:2010:SL:1841234}
\bibfield{author}{\bibinfo{person}{Olivier Chapelle}, \bibinfo{person}{Bernhard
  Schlkopf}, {and} \bibinfo{person}{Alexander Zien}.}
  \bibinfo{year}{2010}\natexlab{}.
\newblock \bibinfo{booktitle}{\emph{Semi-Supervised Learning}
  (\bibinfo{edition}{1st} ed.)}.
\newblock \bibinfo{publisher}{The MIT Press}.
\newblock
\showISBNx{0262514125, 9780262514125}


\bibitem[\protect\citeauthoryear{Charuvaka and Rangwala}{Charuvaka and
  Rangwala}{2015}]%
        {CharuvakaR15}
\bibfield{author}{\bibinfo{person}{Anveshi Charuvaka} {and}
  \bibinfo{person}{Huzefa Rangwala}.} \bibinfo{year}{2015}\natexlab{}.
\newblock \showarticletitle{HierCost: Improving Large Scale Hierarchical
  Classification with Cost Sensitive Learning}. In
  \bibinfo{booktitle}{\emph{{ECML/PKDD} {(1)}}} \emph{(\bibinfo{series}{Lecture
  Notes in Computer Science})}, Vol.~\bibinfo{volume}{9284}.
  \bibinfo{publisher}{Springer}, \bibinfo{pages}{675--690}.
\newblock


\bibitem[\protect\citeauthoryear{Collins}{Collins}{2002}]%
        {DBLP:conf/emnlp/Collins02}
\bibfield{author}{\bibinfo{person}{Michael Collins}.}
  \bibinfo{year}{2002}\natexlab{}.
\newblock \showarticletitle{Discriminative Training Methods for Hidden Markov
  Models: Theory and Experiments with Perceptron Algorithms}. In
  \bibinfo{booktitle}{\emph{{EMNLP}}}.
\newblock


\bibitem[\protect\citeauthoryear{Cong, Lee, Wu, and Liu}{Cong
  et~al\mbox{.}}{2004}]%
        {DBLP:conf/dasfaa/CongLWL04}
\bibfield{author}{\bibinfo{person}{Gao Cong}, \bibinfo{person}{Wee~Sun Lee},
  \bibinfo{person}{Haoran Wu}, {and} \bibinfo{person}{Bing Liu}.}
  \bibinfo{year}{2004}\natexlab{}.
\newblock \showarticletitle{Semi-supervised Text Classification Using
  Partitioned {EM}}. In \bibinfo{booktitle}{\emph{{DASFAA}}}
  \emph{(\bibinfo{series}{Lecture Notes in Computer Science})},
  Vol.~\bibinfo{volume}{2973}. \bibinfo{publisher}{Springer},
  \bibinfo{pages}{482--493}.
\newblock


\bibitem[\protect\citeauthoryear{Dagan, Karov, and Roth}{Dagan
  et~al\mbox{.}}{1997}]%
        {DaganKR97}
\bibfield{author}{\bibinfo{person}{Ido Dagan}, \bibinfo{person}{Yael Karov},
  {and} \bibinfo{person}{Dan Roth}.} \bibinfo{year}{1997}\natexlab{}.
\newblock \showarticletitle{Mistake-Driven Learning in Text Categorization}. In
  \bibinfo{booktitle}{\emph{{EMNLP}}}. \bibinfo{publisher}{{ACL}}.
\newblock


\bibitem[\protect\citeauthoryear{Dalvi, Mishra, and Cohen}{Dalvi
  et~al\mbox{.}}{2016}]%
        {DBLP:conf/wsdm/DalviMC16}
\bibfield{author}{\bibinfo{person}{Bhavana Dalvi},
  \bibinfo{person}{Aditya~Kumar Mishra}, {and} \bibinfo{person}{William~W.
  Cohen}.} \bibinfo{year}{2016}\natexlab{}.
\newblock \showarticletitle{Hierarchical Semi-supervised Classification with
  Incomplete Class Hierarchies}. In \bibinfo{booktitle}{\emph{{WSDM}}}.
  \bibinfo{publisher}{{ACM}}, \bibinfo{pages}{193--202}.
\newblock


\bibitem[\protect\citeauthoryear{Dempster, Laird, and Rubin}{Dempster
  et~al\mbox{.}}{1977}]%
        {dempster1977maximum}
\bibfield{author}{\bibinfo{person}{Arthur~P Dempster}, \bibinfo{person}{Nan~M
  Laird}, {and} \bibinfo{person}{Donald~B Rubin}.}
  \bibinfo{year}{1977}\natexlab{}.
\newblock \showarticletitle{Maximum likelihood from incomplete data via the EM
  algorithm}.
\newblock \bibinfo{journal}{\emph{Journal of the royal statistical society.
  Series B (methodological)}} (\bibinfo{year}{1977}), \bibinfo{pages}{1--38}.
\newblock


\bibitem[\protect\citeauthoryear{Duda and Hart}{Duda and Hart}{1973}]%
        {DudaHart73}
\bibfield{author}{\bibinfo{person}{Richard~O Duda} {and}
  \bibinfo{person}{Peter~E Hart}.} \bibinfo{year}{1973}\natexlab{}.
\newblock \showarticletitle{Pattern classification and scene analysis}.
\newblock \bibinfo{journal}{\emph{A Wiley-Interscience Publication, New York:
  Wiley, 1973}} (\bibinfo{year}{1973}).
\newblock


\bibitem[\protect\citeauthoryear{Dumais and Chen}{Dumais and Chen}{2000}]%
        {DBLP:conf/sigir/DumaisC00}
\bibfield{author}{\bibinfo{person}{Susan~T. Dumais} {and} \bibinfo{person}{Hao
  Chen}.} \bibinfo{year}{2000}\natexlab{}.
\newblock \showarticletitle{Hierarchical classification of Web content}. In
  \bibinfo{booktitle}{\emph{{SIGIR}}}. \bibinfo{publisher}{{ACM}},
  \bibinfo{pages}{256--263}.
\newblock


\bibitem[\protect\citeauthoryear{Fan, Chang, Hsieh, Wang, and Lin}{Fan
  et~al\mbox{.}}{2008}]%
        {fan2008liblinear}
\bibfield{author}{\bibinfo{person}{Rong{-}En Fan}, \bibinfo{person}{Kai{-}Wei
  Chang}, \bibinfo{person}{Cho{-}Jui Hsieh}, \bibinfo{person}{Xiang{-}Rui
  Wang}, {and} \bibinfo{person}{Chih{-}Jen Lin}.}
  \bibinfo{year}{2008}\natexlab{}.
\newblock \showarticletitle{{LIBLINEAR:} {A} Library for Large Linear
  Classification}.
\newblock \bibinfo{journal}{\emph{Journal of Machine Learning Research}}
  \bibinfo{volume}{9} (\bibinfo{year}{2008}), \bibinfo{pages}{1871--1874}.
\newblock


\bibitem[\protect\citeauthoryear{Fox{-}Roberts and Rosten}{Fox{-}Roberts and
  Rosten}{2014}]%
        {DBLP:journals/jmlr/Fox-RobertsR14}
\bibfield{author}{\bibinfo{person}{Patrick Fox{-}Roberts} {and}
  \bibinfo{person}{Edward Rosten}.} \bibinfo{year}{2014}\natexlab{}.
\newblock \showarticletitle{Unbiased generative semi-supervised learning}.
\newblock \bibinfo{journal}{\emph{Journal of Machine Learning Research}}
  \bibinfo{volume}{15}, \bibinfo{number}{1} (\bibinfo{year}{2014}),
  \bibinfo{pages}{367--443}.
\newblock


\bibitem[\protect\citeauthoryear{Gopal and Yang}{Gopal and Yang}{2013}]%
        {GopalY13}
\bibfield{author}{\bibinfo{person}{Siddharth Gopal} {and}
  \bibinfo{person}{Yiming Yang}.} \bibinfo{year}{2013}\natexlab{}.
\newblock \showarticletitle{Recursive regularization for large-scale
  classification with hierarchical and graphical dependencies}. In
  \bibinfo{booktitle}{\emph{{KDD}}}. \bibinfo{publisher}{{ACM}},
  \bibinfo{pages}{257--265}.
\newblock


\bibitem[\protect\citeauthoryear{Gopal, Yang, Bai, and Niculescu{-}Mizil}{Gopal
  et~al\mbox{.}}{2012}]%
        {GopalYBN12}
\bibfield{author}{\bibinfo{person}{Siddharth Gopal}, \bibinfo{person}{Yiming
  Yang}, \bibinfo{person}{Bing Bai}, {and} \bibinfo{person}{Alexandru
  Niculescu{-}Mizil}.} \bibinfo{year}{2012}\natexlab{}.
\newblock \showarticletitle{Bayesian models for Large-scale Hierarchical
  Classification}. In \bibinfo{booktitle}{\emph{{NIPS}}}.
  \bibinfo{pages}{2420--2428}.
\newblock


\bibitem[\protect\citeauthoryear{Koller and Sahami}{Koller and Sahami}{1997}]%
        {DBLP:conf/icml/KollerS97}
\bibfield{author}{\bibinfo{person}{Daphne Koller} {and} \bibinfo{person}{Mehran
  Sahami}.} \bibinfo{year}{1997}\natexlab{}.
\newblock \showarticletitle{Hierarchically Classifying Documents Using Very Few
  Words}. In \bibinfo{booktitle}{\emph{{ICML}}}. \bibinfo{publisher}{Morgan
  Kaufmann}, \bibinfo{pages}{170--178}.
\newblock


\bibitem[\protect\citeauthoryear{Lafferty, McCallum, and Pereira}{Lafferty
  et~al\mbox{.}}{2001}]%
        {LaffertyMP01}
\bibfield{author}{\bibinfo{person}{John~D. Lafferty}, \bibinfo{person}{Andrew
  McCallum}, {and} \bibinfo{person}{Fernando C.~N. Pereira}.}
  \bibinfo{year}{2001}\natexlab{}.
\newblock \showarticletitle{Conditional Random Fields: Probabilistic Models for
  Segmenting and Labeling Sequence Data}. In \bibinfo{booktitle}{\emph{ICML}}.
  \bibinfo{pages}{282--289}.
\newblock


\bibitem[\protect\citeauthoryear{Lang}{Lang}{1995}]%
        {Lang95}
\bibfield{author}{\bibinfo{person}{Ken Lang}.} \bibinfo{year}{1995}\natexlab{}.
\newblock \showarticletitle{NewsWeeder: Learning to Filter Netnews}. In
  \bibinfo{booktitle}{\emph{{ICML}}}. \bibinfo{publisher}{Morgan Kaufmann},
  \bibinfo{pages}{331--339}.
\newblock


\bibitem[\protect\citeauthoryear{Lewis, Yang, Rose, and Li}{Lewis
  et~al\mbox{.}}{2004}]%
        {lewis2004rcv1}
\bibfield{author}{\bibinfo{person}{David~D. Lewis}, \bibinfo{person}{Yiming
  Yang}, \bibinfo{person}{Tony~G. Rose}, {and} \bibinfo{person}{Fan Li}.}
  \bibinfo{year}{2004}\natexlab{}.
\newblock \showarticletitle{{RCV1:} {A} New Benchmark Collection for Text
  Categorization Research}.
\newblock \bibinfo{journal}{\emph{Journal of Machine Learning Research}}
  \bibinfo{volume}{5} (\bibinfo{year}{2004}), \bibinfo{pages}{361--397}.
\newblock


\bibitem[\protect\citeauthoryear{Liu and Yang}{Liu and Yang}{2012}]%
        {liu2012improvement}
\bibfield{author}{\bibinfo{person}{Mingyoug Liu} {and}
  \bibinfo{person}{Jiangang Yang}.} \bibinfo{year}{2012}\natexlab{}.
\newblock \showarticletitle{An improvement of TFIDF weighting in text
  categorization}.
\newblock \bibinfo{journal}{\emph{International proceedings of computer science
  and information technology}} (\bibinfo{year}{2012}), \bibinfo{pages}{44--47}.
\newblock


\bibitem[\protect\citeauthoryear{Liu, Yang, Wan, Zeng, Chen, and Ma}{Liu
  et~al\mbox{.}}{2005}]%
        {DBLP:journals/sigkdd/LiuYWZCM05}
\bibfield{author}{\bibinfo{person}{Tie{-}Yan Liu}, \bibinfo{person}{Yiming
  Yang}, \bibinfo{person}{Hao Wan}, \bibinfo{person}{Hua{-}Jun Zeng},
  \bibinfo{person}{Zheng Chen}, {and} \bibinfo{person}{Wei{-}Ying Ma}.}
  \bibinfo{year}{2005}\natexlab{}.
\newblock \showarticletitle{Support vector machines classification with a very
  large-scale taxonomy}.
\newblock \bibinfo{journal}{\emph{{SIGKDD} Explorations}} \bibinfo{volume}{7},
  \bibinfo{number}{1} (\bibinfo{year}{2005}), \bibinfo{pages}{36--43}.
\newblock


\bibitem[\protect\citeauthoryear{Mann and McCallum}{Mann and McCallum}{2008}]%
        {MannM08}
\bibfield{author}{\bibinfo{person}{Gideon~S. Mann} {and}
  \bibinfo{person}{Andrew McCallum}.} \bibinfo{year}{2008}\natexlab{}.
\newblock \showarticletitle{Generalized Expectation Criteria for
  Semi-Supervised Learning of Conditional Random Fields}. In
  \bibinfo{booktitle}{\emph{{ACL}}}. \bibinfo{publisher}{The Association for
  Computer Linguistics}, \bibinfo{pages}{870--878}.
\newblock


\bibitem[\protect\citeauthoryear{McCallum, Rosenfeld, Mitchell, and
  Ng}{McCallum et~al\mbox{.}}{1998}]%
        {DBLP:conf/icml/McCallumRMN98}
\bibfield{author}{\bibinfo{person}{Andrew McCallum}, \bibinfo{person}{Ronald
  Rosenfeld}, \bibinfo{person}{Tom~M. Mitchell}, {and}
  \bibinfo{person}{Andrew~Y. Ng}.} \bibinfo{year}{1998}\natexlab{}.
\newblock \showarticletitle{Improving Text Classification by Shrinkage in a
  Hierarchy of Classes}. In \bibinfo{booktitle}{\emph{{ICML}}}.
  \bibinfo{publisher}{Morgan Kaufmann}, \bibinfo{pages}{359--367}.
\newblock


\bibitem[\protect\citeauthoryear{Mccord and Chuah}{Mccord and Chuah}{2011}]%
        {McCordC11}
\bibfield{author}{\bibinfo{person}{Michael Mccord} {and} \bibinfo{person}{M
  Chuah}.} \bibinfo{year}{2011}\natexlab{}.
\newblock \showarticletitle{Spam detection on twitter using traditional
  classifiers}. In \bibinfo{booktitle}{\emph{international conference on
  Autonomic and trusted computing}}. Springer, \bibinfo{pages}{175--186}.
\newblock


\bibitem[\protect\citeauthoryear{Meng, Shen, Zhang, and Han}{Meng
  et~al\mbox{.}}{2018}]%
        {DBLP:conf/cikm/MengSZ018}
\bibfield{author}{\bibinfo{person}{Yu Meng}, \bibinfo{person}{Jiaming Shen},
  \bibinfo{person}{Chao Zhang}, {and} \bibinfo{person}{Jiawei Han}.}
  \bibinfo{year}{2018}\natexlab{}.
\newblock \showarticletitle{Weakly-Supervised Neural Text Classification}. In
  \bibinfo{booktitle}{\emph{{CIKM}}}. \bibinfo{publisher}{{ACM}},
  \bibinfo{pages}{983--992}.
\newblock


\bibitem[\protect\citeauthoryear{Meng, Shen, Zhang, and Han}{Meng
  et~al\mbox{.}}{2019}]%
        {DBLP:journals/corr/abs-1812-11270}
\bibfield{author}{\bibinfo{person}{Yu Meng}, \bibinfo{person}{Jiaming Shen},
  \bibinfo{person}{Chao Zhang}, {and} \bibinfo{person}{Jiawei Han}.}
  \bibinfo{year}{2019}\natexlab{}.
\newblock \showarticletitle{Weakly-Supervised Hierarchical Text
  Classification}. In \bibinfo{booktitle}{\emph{{AAAI}}}.
  \bibinfo{publisher}{{AAAI} Press}.
\newblock


\bibitem[\protect\citeauthoryear{Ng and Jordan}{Ng and Jordan}{2001}]%
        {NgJ01}
\bibfield{author}{\bibinfo{person}{Andrew~Y. Ng} {and}
  \bibinfo{person}{Michael~I. Jordan}.} \bibinfo{year}{2001}\natexlab{}.
\newblock \showarticletitle{On Discriminative vs. Generative Classifiers: {A}
  comparison of logistic regression and naive Bayes}. In
  \bibinfo{booktitle}{\emph{{NIPS}}}. \bibinfo{publisher}{{MIT} Press},
  \bibinfo{pages}{841--848}.
\newblock


\bibitem[\protect\citeauthoryear{Nigam, McCallum, and Mitchell}{Nigam
  et~al\mbox{.}}{2006}]%
        {nigam2006semi}
\bibfield{author}{\bibinfo{person}{Kamal Nigam}, \bibinfo{person}{Andrew
  McCallum}, {and} \bibinfo{person}{Tom Mitchell}.}
  \bibinfo{year}{2006}\natexlab{}.
\newblock \showarticletitle{Semi-supervised text classification using EM}.
\newblock \bibinfo{journal}{\emph{Semi-Supervised Learning}}
  (\bibinfo{year}{2006}), \bibinfo{pages}{33--56}.
\newblock


\bibitem[\protect\citeauthoryear{Nilsson}{Nilsson}{1965}]%
        {Nilsson65}
\bibfield{author}{\bibinfo{person}{N.~J. Nilsson}.}
  \bibinfo{year}{1965}\natexlab{}.
\newblock \bibinfo{booktitle}{\emph{Learning machines: Foundations of Trainable
  Pattern-Classifying Systems} (\bibinfo{edition}{1st} ed.)}.
\newblock \bibinfo{publisher}{McGraw-Hill}.
\newblock
\showISBNx{9780070465701}


\bibitem[\protect\citeauthoryear{Pang, Lee, and Vaithyanathan}{Pang
  et~al\mbox{.}}{2002}]%
        {PangLV02}
\bibfield{author}{\bibinfo{person}{Bo Pang}, \bibinfo{person}{Lillian Lee},
  {and} \bibinfo{person}{Shivakumar Vaithyanathan}.}
  \bibinfo{year}{2002}\natexlab{}.
\newblock \showarticletitle{Thumbs up? Sentiment Classification using Machine
  Learning Techniques}. In \bibinfo{booktitle}{\emph{{EMNLP}}}.
\newblock


\bibitem[\protect\citeauthoryear{Peng, Li, He, Liu, Bao, Wang, Song, and
  Yang}{Peng et~al\mbox{.}}{2018}]%
        {PengLHLBWS018}
\bibfield{author}{\bibinfo{person}{Hao Peng}, \bibinfo{person}{Jianxin Li},
  \bibinfo{person}{Yu He}, \bibinfo{person}{Yaopeng Liu},
  \bibinfo{person}{Mengjiao Bao}, \bibinfo{person}{Lihong Wang},
  \bibinfo{person}{Yangqiu Song}, {and} \bibinfo{person}{Qiang Yang}.}
  \bibinfo{year}{2018}\natexlab{}.
\newblock \showarticletitle{Large-Scale Hierarchical Text Classification with
  Recursively Regularized Deep Graph-CNN}. In
  \bibinfo{booktitle}{\emph{{WWW}}}. \bibinfo{publisher}{{ACM}},
  \bibinfo{pages}{1063--1072}.
\newblock


\bibitem[\protect\citeauthoryear{Song and Roth}{Song and Roth}{2014}]%
        {song2014dataless}
\bibfield{author}{\bibinfo{person}{Yangqiu Song} {and} \bibinfo{person}{Dan
  Roth}.} \bibinfo{year}{2014}\natexlab{}.
\newblock \showarticletitle{On Dataless Hierarchical Text Classification}. In
  \bibinfo{booktitle}{\emph{{AAAI}}}. \bibinfo{publisher}{{AAAI} Press},
  \bibinfo{pages}{1579--1585}.
\newblock


\bibitem[\protect\citeauthoryear{Sun and Lim}{Sun and Lim}{2001}]%
        {SunL01}
\bibfield{author}{\bibinfo{person}{Aixin Sun} {and} \bibinfo{person}{Ee{-}Peng
  Lim}.} \bibinfo{year}{2001}\natexlab{}.
\newblock \showarticletitle{Hierarchical Text Classification and Evaluation}.
  In \bibinfo{booktitle}{\emph{{ICDM}}}. \bibinfo{publisher}{{IEEE} Computer
  Society}, \bibinfo{pages}{521--528}.
\newblock


\bibitem[\protect\citeauthoryear{Taskar, Guestrin, and Koller}{Taskar
  et~al\mbox{.}}{2003}]%
        {TaskarGK03}
\bibfield{author}{\bibinfo{person}{Benjamin Taskar}, \bibinfo{person}{Carlos
  Guestrin}, {and} \bibinfo{person}{Daphne Koller}.}
  \bibinfo{year}{2003}\natexlab{}.
\newblock \showarticletitle{Max-Margin Markov Networks}. In
  \bibinfo{booktitle}{\emph{NIPS}}. \bibinfo{pages}{25--32}.
\newblock


\bibitem[\protect\citeauthoryear{Tikk and Bir{\'{o}}}{Tikk and
  Bir{\'{o}}}{2004}]%
        {DBLP:journals/ajiips/TikkB04}
\bibfield{author}{\bibinfo{person}{Domonkos Tikk} {and}
  \bibinfo{person}{Gy{\"{o}}rgy Bir{\'{o}}}.} \bibinfo{year}{2004}\natexlab{}.
\newblock \showarticletitle{A hierarchical test categorization approach and its
  application to {FRT} expansion}.
\newblock \bibinfo{journal}{\emph{Austr. J. Intelligent Information Processing
  Systems}} \bibinfo{volume}{8}, \bibinfo{number}{3} (\bibinfo{year}{2004}),
  \bibinfo{pages}{123--131}.
\newblock


\bibitem[\protect\citeauthoryear{Tsochantaridis, Joachims, Hofmann, and
  Altun}{Tsochantaridis et~al\mbox{.}}{2005}]%
        {DBLP:journals/jmlr/TsochantaridisJHA05}
\bibfield{author}{\bibinfo{person}{Ioannis Tsochantaridis},
  \bibinfo{person}{Thorsten Joachims}, \bibinfo{person}{Thomas Hofmann}, {and}
  \bibinfo{person}{Yasemin Altun}.} \bibinfo{year}{2005}\natexlab{}.
\newblock \showarticletitle{Large Margin Methods for Structured and
  Interdependent Output Variables}.
\newblock \bibinfo{journal}{\emph{Journal of Machine Learning Research}}
  \bibinfo{volume}{6} (\bibinfo{year}{2005}), \bibinfo{pages}{1453--1484}.
\newblock


\bibitem[\protect\citeauthoryear{Wu, Tygert, and LeCun}{Wu
  et~al\mbox{.}}{2017}]%
        {DBLP:journals/corr/abs-1709-01062}
\bibfield{author}{\bibinfo{person}{Cinna Wu}, \bibinfo{person}{Mark Tygert},
  {and} \bibinfo{person}{Yann LeCun}.} \bibinfo{year}{2017}\natexlab{}.
\newblock \showarticletitle{Hierarchical loss for classification}.
\newblock \bibinfo{journal}{\emph{CoRR}}  \bibinfo{volume}{abs/1709.01062}
  (\bibinfo{year}{2017}).
\newblock


\bibitem[\protect\citeauthoryear{Xiao, Zhou, and Wu}{Xiao
  et~al\mbox{.}}{2011}]%
        {DBLP:conf/icml/XiaoZW11}
\bibfield{author}{\bibinfo{person}{Lin Xiao}, \bibinfo{person}{Dengyong Zhou},
  {and} \bibinfo{person}{Mingrui Wu}.} \bibinfo{year}{2011}\natexlab{}.
\newblock \showarticletitle{Hierarchical Classification via Orthogonal
  Transfer}. In \bibinfo{booktitle}{\emph{{ICML}}}.
  \bibinfo{publisher}{Omnipress}, \bibinfo{pages}{801--808}.
\newblock


\bibitem[\protect\citeauthoryear{Yang}{Yang}{1999}]%
        {DBLP:journals/ir/Yang99}
\bibfield{author}{\bibinfo{person}{Yiming Yang}.}
  \bibinfo{year}{1999}\natexlab{}.
\newblock \showarticletitle{An Evaluation of Statistical Approaches to Text
  Categorization}.
\newblock \bibinfo{journal}{\emph{Inf. Retr.}} \bibinfo{volume}{1},
  \bibinfo{number}{1-2} (\bibinfo{year}{1999}), \bibinfo{pages}{69--90}.
\newblock


\end{thebibliography}

\end{document}